\pdfoutput=1
\documentclass[11pt]{article}

\usepackage[]{emnlp2022}

\usepackage{times}
\usepackage{latexsym}

\usepackage[T1]{fontenc}
\usepackage[utf8]{inputenc}

\usepackage{microtype}

\usepackage{booktabs}
\usepackage{graphicx}
\usepackage{subcaption}
\usepackage{multirow}
\usepackage{blindtext}
\usepackage{float} % here for H placement parameter
\usepackage{hyperref}
\usepackage{textcomp, gensymb} % for \degree
\usepackage{enumitem}
\usepackage{amsmath}
\title{Decoding a Neural Retriever's Latent Space for Query Suggestion}

\author{
    Leonard Adolphs$^\dagger$ \And
    Michelle Chen Huebscher$^\ddagger$ \And
    Christian Buck$^\ddagger$ \AND
    Sertan Girgin$^\ddagger$ \And
    Olivier Bachem$^\ddagger$ \And
    Massimiliano Ciaramita$^\ddagger$ \And
    Thomas Hofmann$^\dagger$ \\
    \AND
    \normalfont{$^\dagger$ETH Zürich} \\ \normalfont{\texttt{\small{ladolphs@inf.ethz.ch}}} \And  \normalfont{$^\ddagger$Google Research}
}

\begin{document}
\maketitle
\renewcommand{\arraystretch}{1.2}
\begin{abstract}
Neural retrieval models have superseded classic bag-of-words methods such as BM25 as the retrieval framework of choice. However, neural systems lack the interpretability of bag-of-words models; it is not trivial to connect a query change to a change in the latent space that ultimately determines the retrieval results. To shed light on this embedding space, we learn a ``query decoder'' that, given a latent representation of a neural search engine, generates the corresponding query. We show that it is possible to decode a meaningful query from its latent representation and, when moving in the right direction in latent space, to decode a query that retrieves the relevant paragraph. In particular, the query decoder can be useful to understand ``what should have been asked'' to retrieve a particular paragraph from the collection. We employ the query decoder to generate a large synthetic dataset of query reformulations for MSMarco, leading to improved retrieval performance. On this data, we train a pseudo-relevance feedback (PRF) T5 model for the application of query suggestion that outperforms both query reformulation and PRF information retrieval baselines.
\end{abstract}
\section{Introduction}
Neural encoder models \citep{karpukhin-etal-2020-dense, gtr, contriever} have improved document retrieval in various settings. They have become an essential building block for applications in open-domain question answering \citep{karpukhin-etal-2020-dense, RAG, izacard-grave-2021-leveraging}, open-domain conversational agents \citep{shuster-etal-2021-retrieval-augmentation, k2r}, and, recently, language modeling \citep{seeker}.
Neural encoders embed documents and queries in a shared (or joint) latent space, so that  paragraphs can be ranked and retrieved based on their vector similarity with a given query. This constitutes a conceptually powerful approach to discovering semantic similarities between queries and documents that is often found to be more nuanced than simple term frequency statistics typical of classic sparse representations. %bag-of-words procedures.
However, such encoders may come with shortcomings in practice. First, they are prone to domain overfitting, failing to consistently outperform bag-of-words approaches on out-of-domain queries \citep{BEIR}. Second, they are notoriously hard to interpret as similarity is no longer controlled by word overlap, but rather by semantic similarities that lack explainability. Third, they may be non-robust as small changes in the query can lead to  inexplicably different retrieval results.

\begin{figure*}
    \centering
    \begin{subfigure}[c]{.21\textwidth}
        \centering
        \includegraphics[width=.9\linewidth]{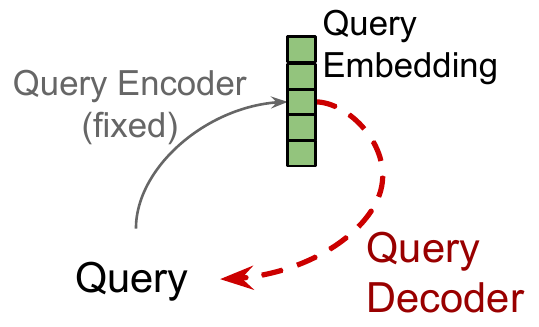}
        \caption{}
        \label{fig:query_decoder_overview}
    \end{subfigure}
    \begin{subfigure}[c]{.32\textwidth}
        \centering
        \includegraphics[width=1.\linewidth]{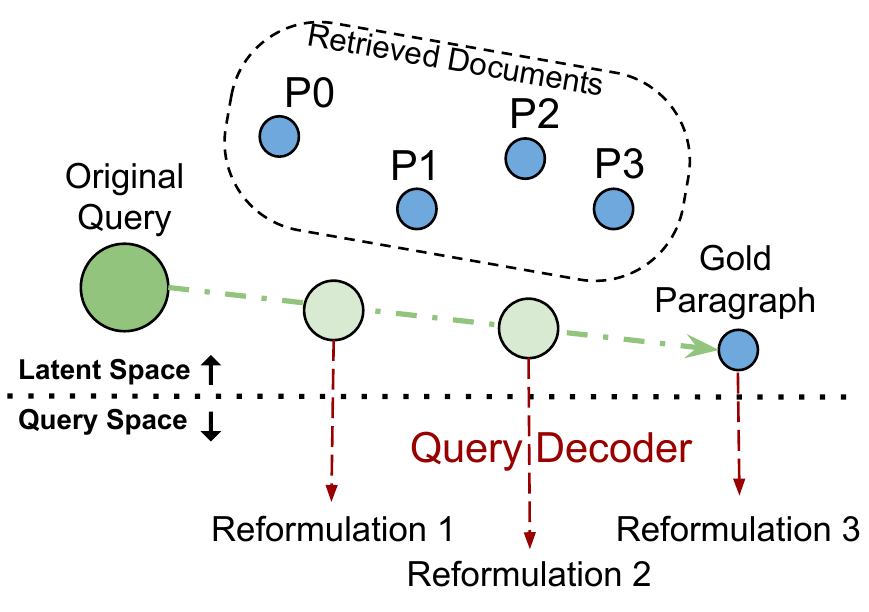}
        \caption{}
        \label{fig:traversal}
    \end{subfigure}
    \begin{subfigure}[c]{.45\textwidth}
        \centering
        \includegraphics[width=1.\linewidth]{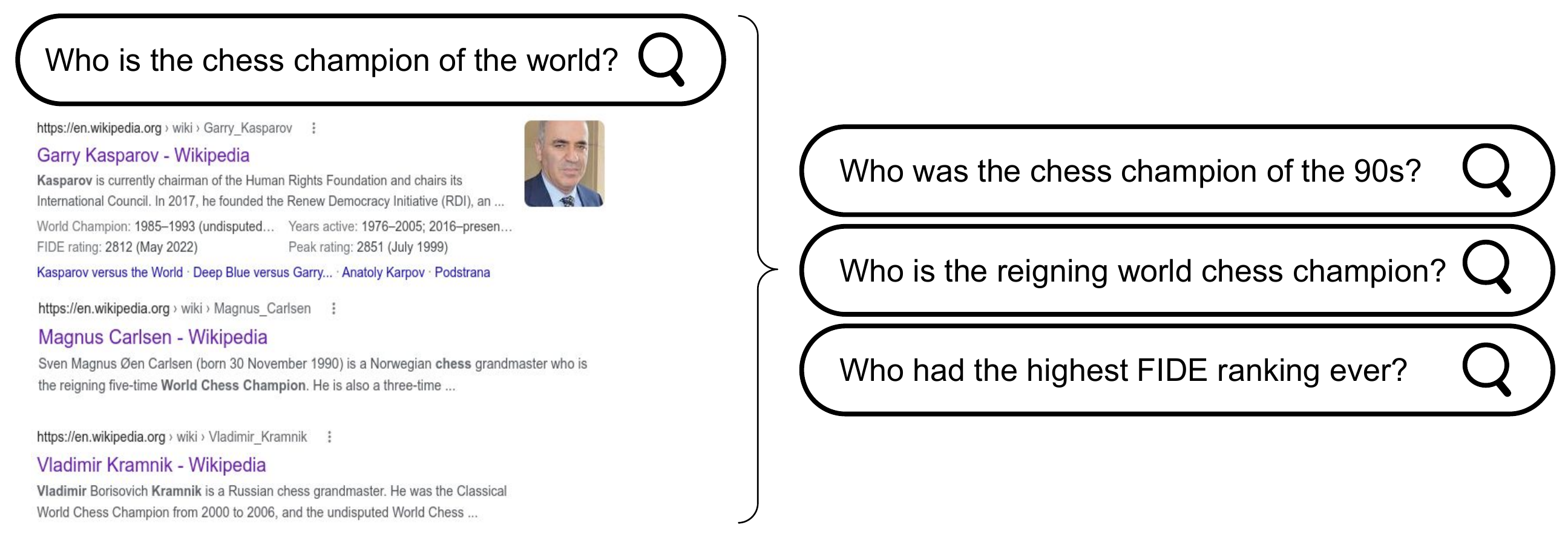}
        \caption{}
        \label{fig:query_suggestion}
    \end{subfigure}
    \label{fig:page_1_img}
    \caption{We train a query decoder (QD) model that inverts the shared encoder of a neural retrieval model (a). Then, we leverage the structure of the latent space of a neural retrieval model by traversing from query to gold paragraph embeddings and using our query decoder to generate a dataset of successful query reformulations (b). Finally, we train a pseudo-relevance feedback query suggestion model on this dataset that predicts promising rewrites, given a query and its search results (c).}
\end{figure*}

In bag-of-words models, it can be straightforward to modify a query to retrieve a given document: e.g., following insights from \emph{relevance feedback}~\citep{rocchio71relevance}, by increasing the weight %relevancy of high tf-idf scoring 
of terms contained in the target document~\citep{ciaramitaboosting,hare}. This approach is not trivially applicable to neural retrieval models as it is unclear how an added term might change the latent code of a query.

In this paper, we look into the missing link connecting latent codes back to actual queries. We thus propose to train a ``query decoder'', which maps  embeddings in the shared query-document space to query strings, inverting the fixed encoder of the neural retriever (cf.~Figure \ref{fig:query_decoder_overview}). As we will show, such a decoder lets us find queries that are optimized to retrieve a given target document. It deciphers what information is in the latent code of a document and how to phrase a query to retrieve it.

We use this model to explore the latent space of a state-of-the-art neural retrieval model, GTR~\citep{gtr}. In particular, we leverage the structure of the latent space by traversing from the embedding of a specific query to its human-labeled gold paragraph and use our query decoder to generate reformulation examples from intermediate points along the path as shown in Figure \ref{fig:traversal}. We find that using this approach, we can generate a large dataset of query reformulations on MSMarco-train  \citep{msmarco} that improve retrieval performance without needing additional human labeling. 
We use this dataset to train a pseudo-relevance feedback (PRF) query suggestion model. 
Here, we fine-tune a T5-large model \citep{t5} that uses the original query, together with its top-5 GTR search results, as the input context to predict a query suggestions as depicted in Figure \ref{fig:query_suggestion}. We show that our model provides fluent, diverse query suggestions with better retrieval performance than various baselines, including a T5 model trained on question editing \citep{Chu_Chen_Chen_Wang_Gimpel_Faruqui_Si_2020}, and a PRF query expansion model \cite{rm3_query_expansion}.

We make the resources to reproduce the results publicly available\footnote{\href{https://github.com/leox1v/query\_decoder}{https://github.com/leox1v/query\_decoder}}.
\section{Related Work} \label{sec:related-work}

\paragraph{Neural Retriever}
Classic retrieval systems such as BM25 \citep{bm25} use term frequency statistics to determine the relevancy of a document for a given query. Recently, neural retrieval models have become more popular and started to outperform classic systems on multiple search tasks. \citet{karpukhin-etal-2020-dense} use a dual-encoder setup based on BERT-base \citep{devlin-etal-2019-bert}, called DPR, to encode query and documents separately and use maximum inner product search \citep{mips} to find a match. They use this model to improve recall and answer quality for multiple open-domain question-answer datasets, including OpenQA-NQ \citep{lee-etal-2019-latent}. \citet{gtr} show that scaling up the dual encoder architecture improves the retrieval performance. They train a shared dual encoder model, based on T5 \citep{t5}, in a multi-stage manner, including fine-tuning on MSMarco \citep{msmarco}, and evaluate on the range of retrieval tasks of the BEIR benchmark \citep{BEIR}. \citet{contriever} show that one can train an unsupervised dense retriever and be competitive against strong baselines on the BEIR benchmark.

\citet{ance} propose approximate
nearest neighbor negative contrastive learning (ANCE) to learn a dense retrieval system. On top of this dense retriever, \citet{ance_prf} consider a pseudo-relevance feedback method. Other than our approach, this method does not provide the user with rephrased queries.

\paragraph{Applications of Neural Retrievers}
Neural retrieval models have been at the core of recent improvements among a range of different NLP tasks. \citet{RAG} augment a language generation model, BART \citep{lewis-etal-2020-bart}, with a DPR neural retriever and evaluate on multiple knowledge-intensive NLP tasks; most notably, they improve over previous models on multiple open-domain QA benchmarks using an abstractive method.

\citet{izacard-grave-2021-leveraging} propose the Fusion-in-Decoder method to aggregate a large set of documents from the neural retriever and provide them to the model during answer generation. Their focus is on open-domain QA where they significantly outperform previous models when considering a large set of documents during decoding.

\citet{shuster-etal-2021-retrieval-augmentation} use neural retrieval models to improve conversational agents in knowledge-grounded dialogue. They show that the issue of hallucination -- i.e., generating factual incorrect knowledge statements -- can be significantly reduced when using a neural-retriever-in-the-loop architecture. Separating the retrieval-augmented knowledge generation and the conversational response generation can further improve the issue of hallucination in knowledge-grounded dialogue and helps fuse modular QA and dialogue models \citep{k2r}. Recently, retrieval query generation approaches have been proposed to improve open-domain dialogue \citep{internet_augmented_dialogue_generation} and language modeling \citep{seeker}.

\paragraph{Query Generation}
Query optimization is a long-standing problem in IR~\citep{lau-horowitz,teevan-04}. Recent work has investigated query refinement with reinforcement learning for Open Domain and Conversational Question Answering~\citep{nogueira-cho-2017-task,buck2018ask,conqrr}.

The methods presented in this paper are a natural complement to the work of
\citet{ciaramitaboosting}, who propose a heuristic approach to generate multi-step query refinements, used to train sequential query generation models for the task of \emph{learning to search}. Their method is also inspired by relevance feedback, but they seek to reach the gold document purely in language space, by brute force exploration. For this purpose, they use specialized search operators to condition the retrieval results as desired. \citet{hare} show that, when paired with a hybrid sparse/dense retrieval environment, the search agents trained on this kind of synthetic data combine effective corpus exploration, competitive performance and interpretability. 

Web-GPT~\citep{DBLP:journals/corr/abs-2112-09332} presents an end-to-end search modeling approach based on human demonstrations, in a similar spirit our work could be seen as way of involving humans-in-loop by proposing better queries.

\paragraph{Fixed-vector decoders}
Probabilistic decoders mapping from a fixed size vector space to natural language have also been explored in auto-encoder settings. A key challenge in this line of work lies in obtaining decoders that are robust, \emph{i.e.}, they generate natural text for a variety of input vectors. \citet{bowman-etal-2016-generating} proposed using a RNN-based language model in combination with variational autoencoders (VAE) \citep{kingma2013auto} which add Gaussian space to the decoder input. \citet{zhao2018adversarially} proposed the use of Adversarial Autoencoder (AAE) \citep{makhzani2015adversarial} to which \citet{pmlr-v119-shen20c} added data denoising by randomly dropping words in the input and the reconstructing the full output.

Recently, RNN-based decoders have been replaced by Transformer-based language models~\citep{vaswani2017attention}, for example by \citet{montero-etal-2021-sentence}, \citet{park2021finetuning} and \citet{li2020optimus}. 
\section{Query Decoder}\label{sec:query_decoder_model}

\begin{table}[]
    \centering
    \includegraphics[width=.95\linewidth]{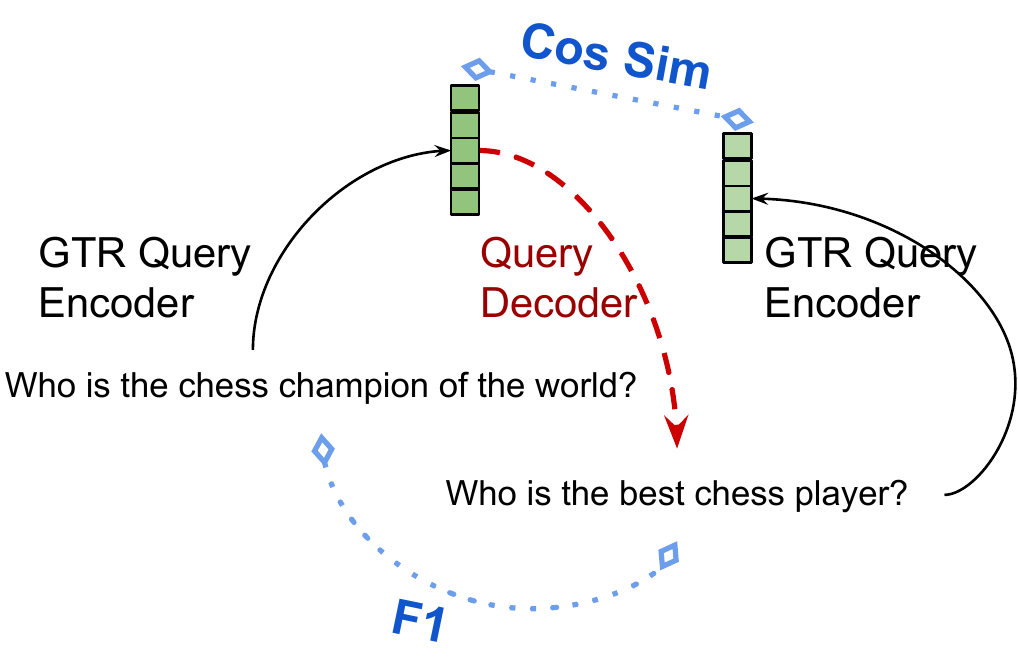}
    \small
    \begin{tabular}{lrr}
        \toprule
          Data &    F1 &  Cos Sim \\
        \midrule
         MSMarco & 0.750 &    0.960 \\
         NQ & 0.886 &    0.980 \\
        \bottomrule
    \end{tabular}
    \caption{Decoding metrics of the Query Decoder (QD) based on the GTR-base neural retrieval model. The F1 score is the F1 word overlap between the original query, of MSMarco or NQ, and the output of the query decoder model when provided with the GTR encoding of the query. The cosine similarity is measured between the re-encoding of the generated query and the encoding of the original query. The figure above depicts the metrics visually with a toy example for clarity.}
    \label{tab:qd_eval}
\end{table}

\paragraph{Training}
We train a T5 \citep{t5}, decoder-only model, to (re-)generate a query from its embedding obtained in a neural retrieval model. As training data, we use a subset of 3 million queries of the PAQ dataset \citep{lewis-etal-2021-paq}. We use the GTR-base \citep{gtr}, shared-encoder model, to generate the embeddings and use the queries as the targets. The objective of the query decoder learning is to invert the mapping of the \textit{fixed} GTR encoder model, as visually depicted in Figure \ref{fig:query_decoder_overview}. More training details of the query decoder are provided in Appendix \ref{sec:qd_details}.

\paragraph{Query Reconstruction Evaluation}
We consider the round-trip consistency as a first step in evaluating the query decoder's effectiveness. A query $q$ is encoded via GTR and then decoded by our decoder to generate  $q'$. We use queries from MSMarco, and NQ test sets of the BEIR benchmark \cite{BEIR}. As a first metric, we compute the F1 score between the original $q$ and its reconstruction $q'$. Since word-overlap is imperfect in measuring query drift, we further re-encode $q'$ and compare its latent code with the code for $q$ via their cosine similarity. The results of these evaluations are reported in Table \ref{tab:qd_eval}, where we also provide an illustrative example of the proposed approach. For both datasets, MSMarco and NQ, the metrics of F1 and cosine similarity are generally high, indicating that the GTR code carries information that allows for close approximate query reconstruction.

\begin{table}[th]
\small
    \centering
    \includegraphics[width=.99\linewidth]{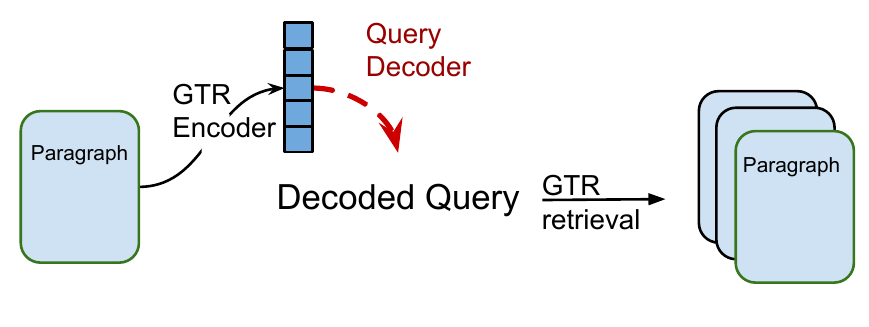}
    \begin{tabular}{lrrr}
        \toprule
           Data &  Top1 &  Top3 &  Top5 \\
        \midrule
           MSMarco & 0.551 & 0.737 & 0.796 \\
           NQ & 0.721 & 0.863 & 0.897 \\
        \bottomrule
    \end{tabular}
    \caption{Share of gold paragraphs for which we can decode a query that retrieves the given paragraph within its top-k GTR search results. The figure above depicts the metric evaluation visually for clarity.}
    \label{tab:qd_paragraph_eval}
\end{table}

\paragraph{Paragraph to Query Evaluation}

Many interesting use cases rely on the ability to generate queries from passages of text \citep{https://doi.org/10.48550/arxiv.1705.00106, DBLP:journals/corr/abs-1803-03664}.
As GTR embeds document paragraphs and queries into the the same space, the query decoder can also be used to invert the retrieval process. We thus evaluate the decoder quality by starting from a document paragraph, decoding a query from its embedding and then running the GTR search engine on that query to check if this query retrieves the desired paragraph as a top-ranked result. We test this in an experiment with human-labeled gold paragraphs from MSMarco and NQ, using top-k as the success metric. The results reported in Table \ref{tab:qd_paragraph_eval} are very encouraging in that the desired paragraph is indeed found very often  among the topmost GTR search results. Two example paragraph decodings from MSMarco are shown in Table \ref{tab:paragraph_decoding_examples}; for both decodings, the gold paragraph is retrieved at the top position.

\renewcommand{\arraystretch}{1.3}
\begin{table}[t]
    \small
    \centering
    \begin{tabular}{p{\linewidth}}
        \toprule
        \textbf{Original Query} \\
        \hspace{0.2cm} nebl coin price  \hfill [Rank: 2]\\ %\hline 
        
        \textbf{Decoding from Gold Paragraph} \\
         \hspace{0.2cm} what is the current price of neblio today belo \hfill [Rank: 1]  \\
       % \hline
        \textbf{Gold Paragraph} \\
        \hspace{0.2cm} Neblio Price Chart US Dollar (NEBL/USD) Neblio price for today is \$16.3125. It has a current circulating supply of 12.8 Million coins and a total volume exchanged of \$9,701,465
 \\
        \midrule
        
        \textbf{Original Query} \\
        \hspace{0.2cm} when is champaign il midterm elections  \hfill [Rank: 3]\\ %\hline 
        \textbf{Decoding from Gold Paragraph} \\
         \hspace{0.2cm} when is the general election in illinois 2018 \hfill [Rank: 1]  \\
      %  \hline
        \textbf{Gold Paragraph} \\
        \hspace{0.2cm} Illinois elections, 2018. A general election will be held in the U.S. state of Illinois on November 6, 2018. All of Illinois' executive officers will be up for election as well as all of Illinois' eighteen seats in the United States House of Representatives.
 \\
  \bottomrule

    \end{tabular}
    \caption{Examples of query decodings from the gold paragraph. The rank indicates the retrieval position of the gold paragraph using the corresponding query.}
    \label{tab:paragraph_decoding_examples}
\end{table}

\begin{figure}[t]
    \centering
    \includegraphics[width=1.0\linewidth]{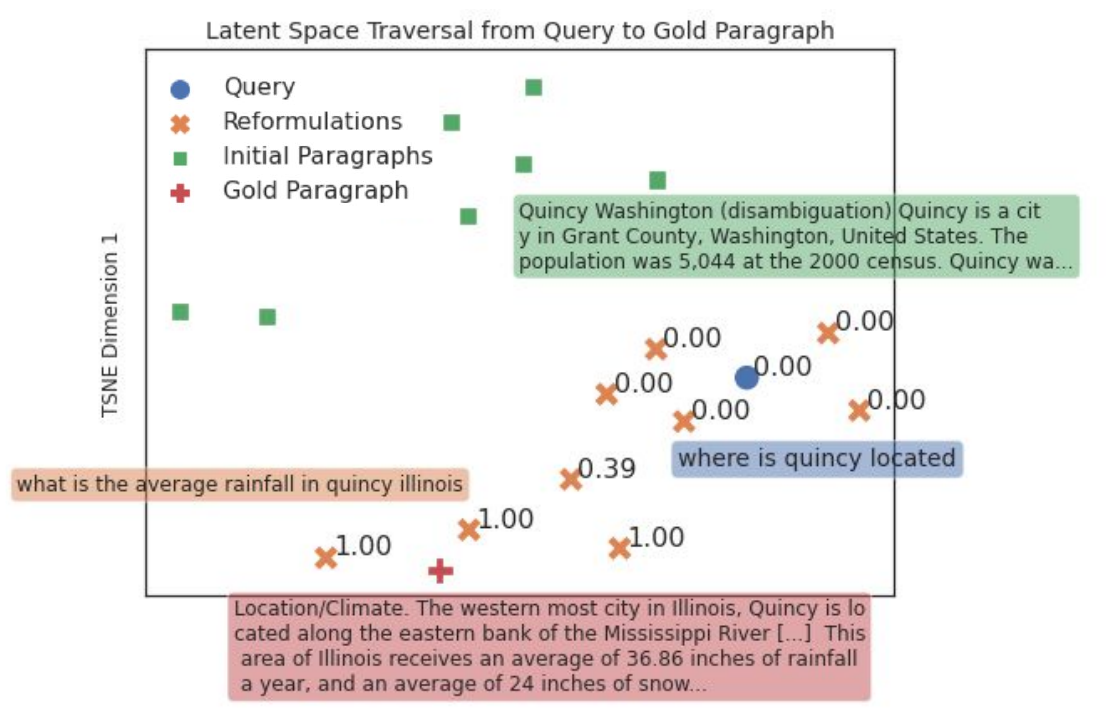}
    \caption{Visualization of the latent-space traversal from query to gold paragraph, using 2D t-SNE. The blue point denotes the embedding of the original query ``where is quincy located''. The green squares are the embeddings of the retrieved paragraphs for this query. The closest one about ``Quincy Washington'' is shown in the green text bar. The orange crosses denote the embeddings of the reformulations of the query decoder when moving to the gold paragraph depicted as the red plus. The orange and red text bars show the final reformulation and the gold paragraph text, respectively. The number above the query and reformulations show the nDCG score. As the gold paragraph is describing the climate of Quincy in addition to its location, a reformulation about the ``average rainfall in quincy illinois'' retrieves the desired paragraph.}
    \label{fig:traversal_visualization}
\end{figure}

\begin{figure}
    \centering
    \begin{subfigure}[b]{.49\textwidth}
        \centering
        \includegraphics[width=1\linewidth]{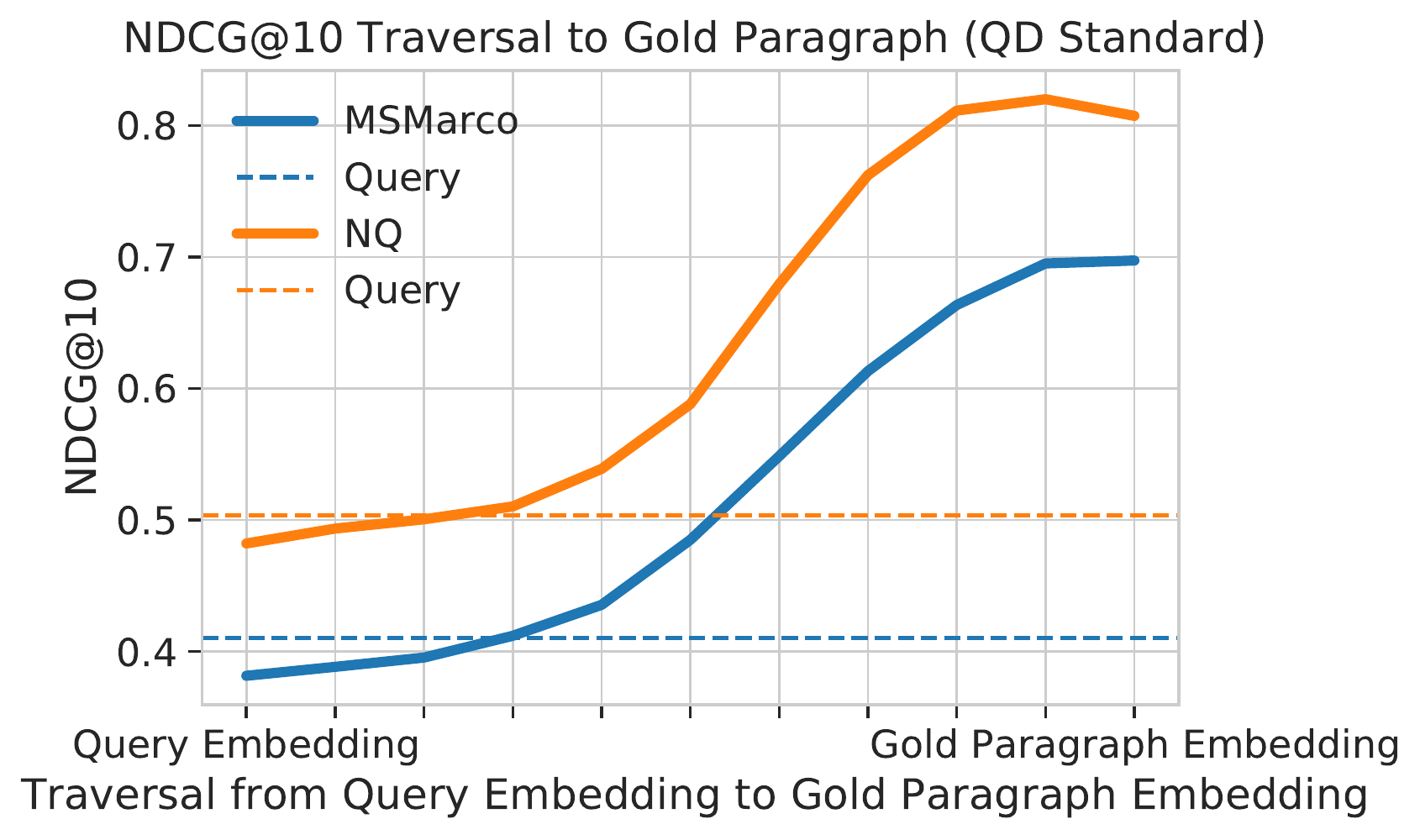}
    \end{subfigure}
    \caption{The normalized discounted cumulative gain (nDCG) of the reformulations from the query decoder when moving the input code from the embedding of the query to the embedding of the gold paragraph. Decoding closer to the gold paragraph embedding leads to queries with improved average retrieval performance. The initial decodings nDCG scores are slightly lower than from the original query due to the reconstruction loss of the query decoder.}
    \label{fig:full_traversal_fig}
\end{figure}

\paragraph{Latent Space Traversal Decoding}
We have shown that query decoding can reconstruct queries and that it can find retrieval queries for target passages. We now turn to a more concrete practical application, namely to automatically generate a data set of query reformulations, from which strategies for interactive retrieval can be learned. In this context, reformulated queries should remain semantically similar to the original query and not overfit to the target passage. They should be somewhat 'in between' the query and the gold passage, as any passage is likely to contain answers to multiple, different questions. This can be operationalized by decoding queries from points along the line connecting the embeddings of the query and its target passage as depicted in Figure \ref{fig:traversal}. 

To validate this idea, we apply it to the MSMarco and NQ retrieval dataset where each query is paired with a human-labeled gold paragraph. In particular, we move in $k$ equidistant increments from the original query embedding $\mathbf{q}$ to the gold paragraph embedding $\mathbf{d}$, i.e. 
\begin{align}
     \mathbf{q}_\kappa = \mathbf{q} + \frac{\kappa}{k} (\mathbf{d} - \mathbf{q}) \qquad \kappa = 0, \dots, k
\end{align}
and generate a reformulation at each step.\footnote{We underline that this procedure can be seen as a latent space equivalent of the 'Rocchio Session' process for generating synthetic search sequences of~\citet{ciaramitaboosting}.} As a sanity check, Figure \ref{fig:full_traversal_fig} shows the average retrieval performance of the decoded queries when moving from the original query embedding to the gold paragraph embedding for MSMarco and NQ. For both datasets, the normalized discounted cumulative gain (nDCG) \citep{10.1145/582415.582418} improves steadily and plateaus, then slightly dips, only when getting close to the gold paragraph embedding. We hypothesize that two effects are at work here that explain this dip: (i) the closer one moves towards the gold passage embedding, the more the query decoder operates out-of-distribution as it is trained on query embeddings. The joint latent space is sparse and likely characterized by distinct regions for queries and passage embeddings, which have different properties (e.g., length or surface structure). (ii) A passage might answer several questions. When decoding from an embedding close to the paragraph, these might start being conflated.

\paragraph{Examples} We provide a visual example of the latent traversal approach in Figure \ref{fig:traversal_visualization} where we project the latent space to 2 dimensions using t-SNE \citep{tsne}. The plot shows that for the ambiguous query ``where is quincy located'' (blue dot), the gold paragraph about the climate of Quincy, Illinois (red plus), is far away from the top-10 retrieved documents (green squares). Traversing the latent space from the query towards the gold paragraph leads to improved reformulations (orange crosses), as is evident from the shrinking distance to the gold paragraph and by the improved nDCG score. 

Semantically, the reformulations move to questions about the climate of Quincy, as this is the main topic of the gold paragraph. The full sequence of reformulations is shown in Table \ref{tab:traversal_visualization_session}. Another example is shown in Table \ref{tab:successful_traversal}; similarly, we see that the decoded queries move semantically from the general question of average annual return on the \textit{stock market} to the more specific question of return at the \textit{S\&P stock exchange}.
Many more examples are provided in Appendix \ref{sec:additional_traversal_examples}.

% Successful Retrieval.
\renewcommand{\arraystretch}{1.4}
\begin{table}[t]
    \small
    \centering
    \begin{tabular}{@{}p{0.95\linewidth}@{}}
        \toprule
        \textbf{Original Query} \\
        \hspace{0.2cm} average yearly return on stock market \hfill [0.00] \\% \hline 
        
        \textbf{Decodings during Traversal} \\
        \hspace{0.2cm} what is the average annual return on stock market \hfill [0.00] \\
          \hspace{0.2cm} average return on a stock market year \hfill [0.00] \\
          \hspace{0.2cm} what is the average annual return on stock market \hfill [0.00] \\
          \hspace{0.2cm} what is the average return on stock in a year \hfill [0.00] \\
          \hspace{0.2cm} what is the average return in a stock market \hfill [0.00] \\
          \hspace{0.2cm} what is the average annual return in stock ( s\&p ) \hfill [0.36] \\
          \hspace{0.2cm} what is the average return on the stock market ( s\&p )  \phantom{mm} \hfill [0.36] \\
          \hspace{0.2cm} what is the average return on the s\&p stock exchange at a time  \hfill [1.00] \\
          \hspace{0.2cm} what is the average return in s\&p stock at a time \hfill [0.36] \\
          \hspace{0.2cm} what is the average annual return of the s\&p stock exchange ( best )  \hfill [1.00] \\
      %  \hline 
        \textbf{Gold Paragraph} \\
        \hspace{0.2cm} The S\&P 500 gauges the performance of the stocks of the 500 largest, most stable companies in the Stock Exchange. It is often considered the most accurate measure of the stock market as a whole. The current average annual return from 1926, the year of the S\&Ps inception, through 2011 is 11.69\%. That's a long look back, and most people aren't interested in what happened in the market 80 years ago. \\ \bottomrule

    \end{tabular}
    \caption{Example of a successful traversal on an MSMarco query. The nDCG@10 score of each query is provided in the brackets. Here, we traverse the latent space in $k=10$ equidistant steps from the embedding of the original query ``average yearly return on stock market'', which results in an nDCG retrieval score of 0, to the embedding of the gold paragraph. The queries decoded from a latent code close to the gold paragraph, focus on the returns of the S\&P (as the gold paragraph) and lead to improved retrieval results.}
    \label{tab:successful_traversal}
\end{table}

\section{Query Suggestion Model} \label{sec:reformulation-model}

\begin{figure}[t]
    \centering
    \begin{subfigure}[b]{.49\textwidth}
        \centering
        \includegraphics[width=1.\linewidth]{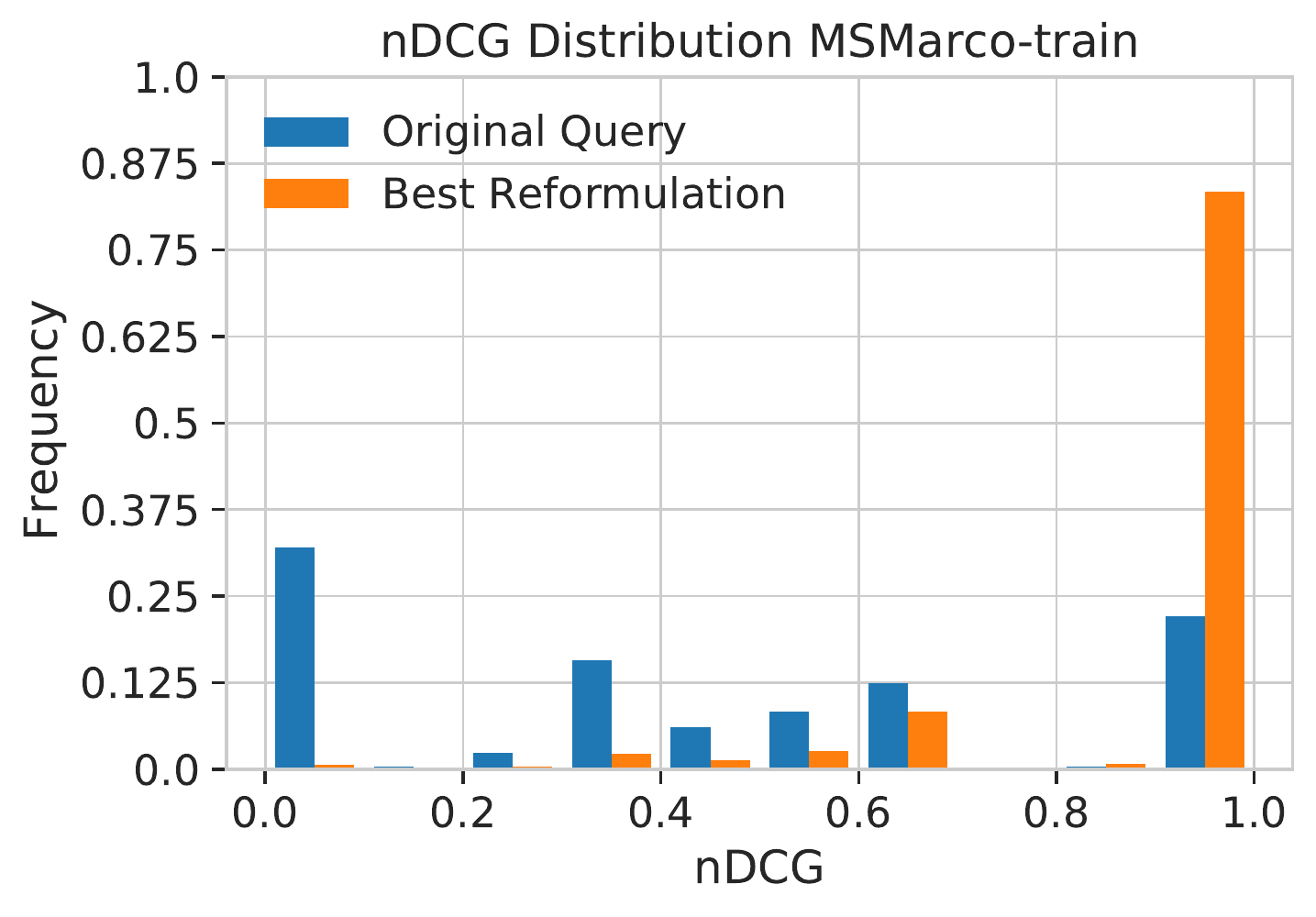}
        \caption{}
        % \caption{The histogram of normalized discounted cumulative gain (nDCG) of the original query vs. the best reformulation when using the latent-space traversal approach on MSMarco-train.}
        \label{fig:ndcg_traversal_train}
    \end{subfigure}
    
    \begin{subfigure}[b]{.49\textwidth}
        \centering
        \includegraphics[width=1.\linewidth]{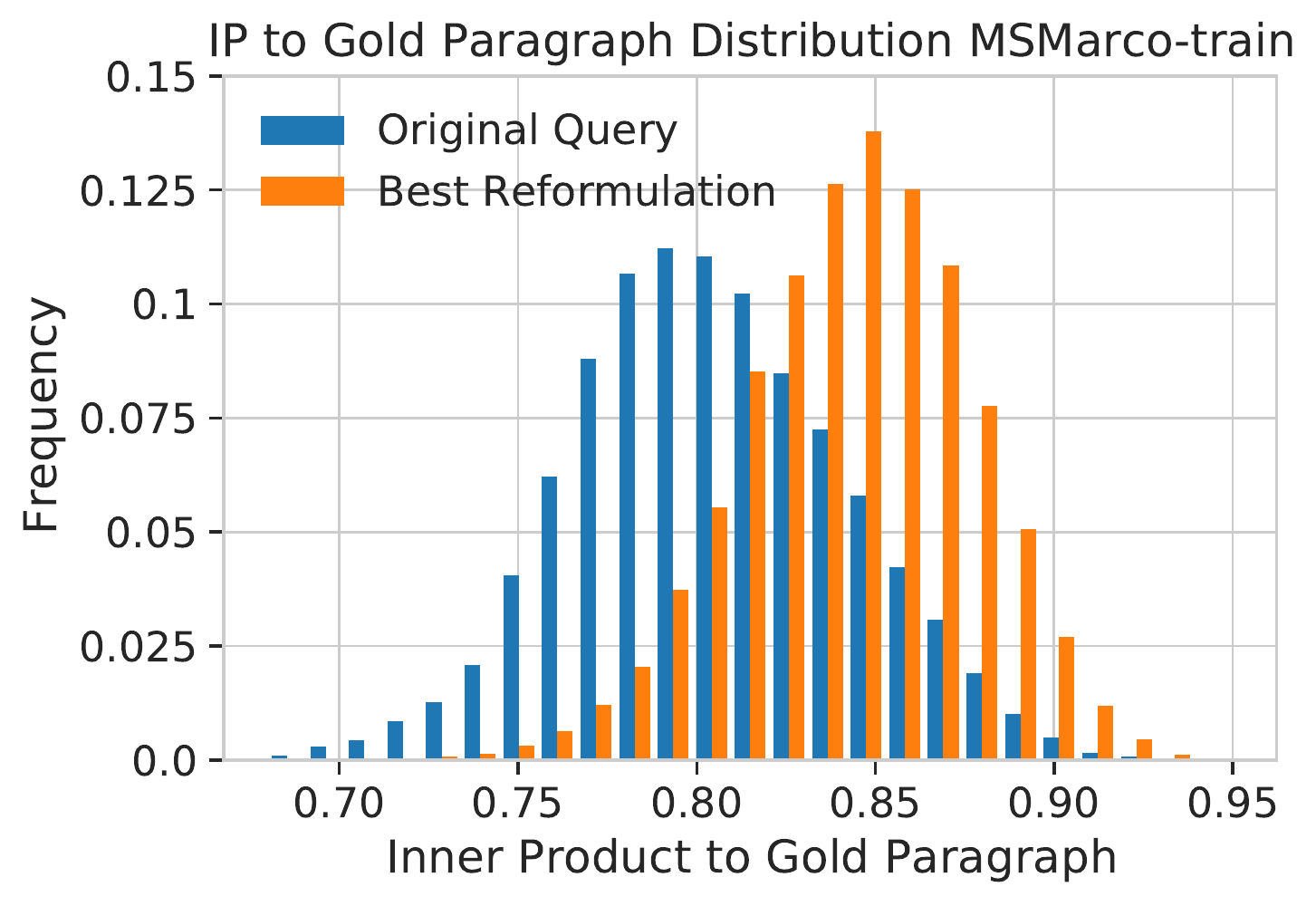}
        \caption{}
        % \caption{The histogram of inner product with the gold paragraph embedding of the original query vs. the best reformulation when using the latent-space traversal approach on MSMarco-train.}
        \label{fig:ip_traversal_train}
    \end{subfigure}
    
    \caption{The histogram of nDCG (a) and inner product with the gold paragraph embedding of the original query vs. the best reformulation found with the latent-space traversal approach on MSMarco-train.}
    \label{fig:full_traversal_train}
\end{figure}

\paragraph{Dataset Generation}
We generate a dataset of query reformulations using the latent space traversal decoding as described in the previous section. In particular, for the 532,761 queries of the MSMarco-train dataset, we leverage GTR's learned latent space structure and move towards the embedding of their gold paragraph. At $k=20$ intermediate steps on this path, we use our query decoder to generate reformulations. 

For more than 80\% of the queries, we find at least one optimal reformulation that retrieves the gold paragraph at the top position.
In Figure \ref{fig:full_traversal_train}, we show histograms of nDCG and the inner product to the gold paragraph for the original query versus the best-found reformulation. The metrics show that the latent space traversal helps us discover good query reformulations that lead to massively improved retrieval performance and are closer to the corresponding gold paragraphs in latent space.

We filter the dataset to only contain ``successful'' reformulations to train the reformulation model. Here, we require a reformulation to have an nDCG of 1 (i.e., retrieve the gold paragraph at the top position), to improve the nDCG compared to the original query, and its embedding to have a larger inner product with the gold paragraph than the original query. Using this approach, we generate a dataset of 863,307 successful query rewrites. As the example in Table \ref{tab:successful_traversal} and Appendix \ref{sec:additional_traversal_examples} show, the decoded queries do not always have human-like fluency and for some sequences intent shift occurs when decoding closer to the paragraph. This is one reason we’re moving in increments from the original query to the gold paragraph instead of directly decoding it.

Interestingly, however, we find that this noise of the dataset is unspecific enough that it gets smoothed out during model training as described in the following paragraph.
More details about this dataset are provided in Sec. \ref{sec:dataset_details}.

\paragraph{Model Training}
We use the reformulation dataset to train two query suggestion model variants. First, we train a model on the plain reformulation examples, from original query to ``successful'' rewrite. As a second, more powerful approach, we train a model with pseudo-relevance feedback (PRF); here, we provide GTR's top-5 search results for the original query as additional context to the model. Both models are fine-tuned from the T5-large \citep{t5} pre-training checkpoint. Consequently, we name the models in the following way:
\begin{itemize}[leftmargin=*,itemsep=2.5pt]
    \item \textbf{qsT5-plain}: A T5 \textbf{q}uery \textbf{s}uggestion model trained on the \textbf{plain} generated reformulation examples (no pseudo-relevance feedback) of MSMarco-train, mapping from query to query reformulation.
    \item \textbf{qsT5}: A T5 \textbf{q}uery \textbf{s}uggestion model trained on the generated reformulation examples of MSMarco-train, where the input is augmented with the content of GTR's top-5 retrieved search results for the original query, mapping from query and search results to query reformulation.
\end{itemize}
The details and hyperparameters of the model training are provided in Appendix \ref{sec:qst5_details}.

\paragraph{Baseline Models}
To measure the effectiveness of our query suggestion model, we benchmark it against multiple baselines. The baselines are meant to cover various angles of competitive approaches to query suggestion, namely training on human-generated question-edit histories, a classic RM3 pseudo-relevance feedback query expansion method, and a latent space sampling approach utilizing our query decoder. In the following, we introduce the three baselines in detail.

\begin{itemize}[leftmargin=*,itemsep=2.5pt]
    \item \textbf{MQR}: We train a T5-large model on the ``multi-domain question rewriting'' (MQR) \citep{Chu_Chen_Chen_Wang_Gimpel_Faruqui_Si_2020} dataset. This dataset consists of 427,719 human-contributed Stack Exchange question-edit histories, mapping from ill-formed to well-formed. While this is a relatively large training dataset, our synthetic generations dataset is roughly double in size with 863,307 rewrites, yet without any human edits.  This baseline captures the effect of turning a query to a well-formed question to improve retrieval performance. It does not use PRF but maps from query to reformulation, as our qsT5-plain model. Training details of this model are provided in Section \ref{sec:mqr_details} in the Appendix.
    \item \textbf{RM3}: We employ RM3 \citep{rm3_novelty_and_hard} as a strong pseudo-relevance feedback baseline. In particular, we use a query expansion approach that uses the formula described in Eq. 20 of \citet{rm3_query_expansion} with $\mu = 2500$ to determine the most relevant terms of the top-5 retrieved documents. Then, each suggestion of the model consists of the original query together with one of the determined relevant terms.
    \item \textbf{Sampling+QD}: To check how much of the retrieval performance gain is due to an ensembling effect in latent space, we compare against a random sampling baseline that includes our query decoder (QD). In particular, we sample a point uniformly at random from an epsilon-ball around the embedding of the original query and use the query decoder to decode that point to a query. This baseline does not use PRF.
\end{itemize}

\begin{figure*}[]
    \centering
    \begin{subfigure}[b]{.49\textwidth}
        \centering
        \includegraphics[width=1.\linewidth]{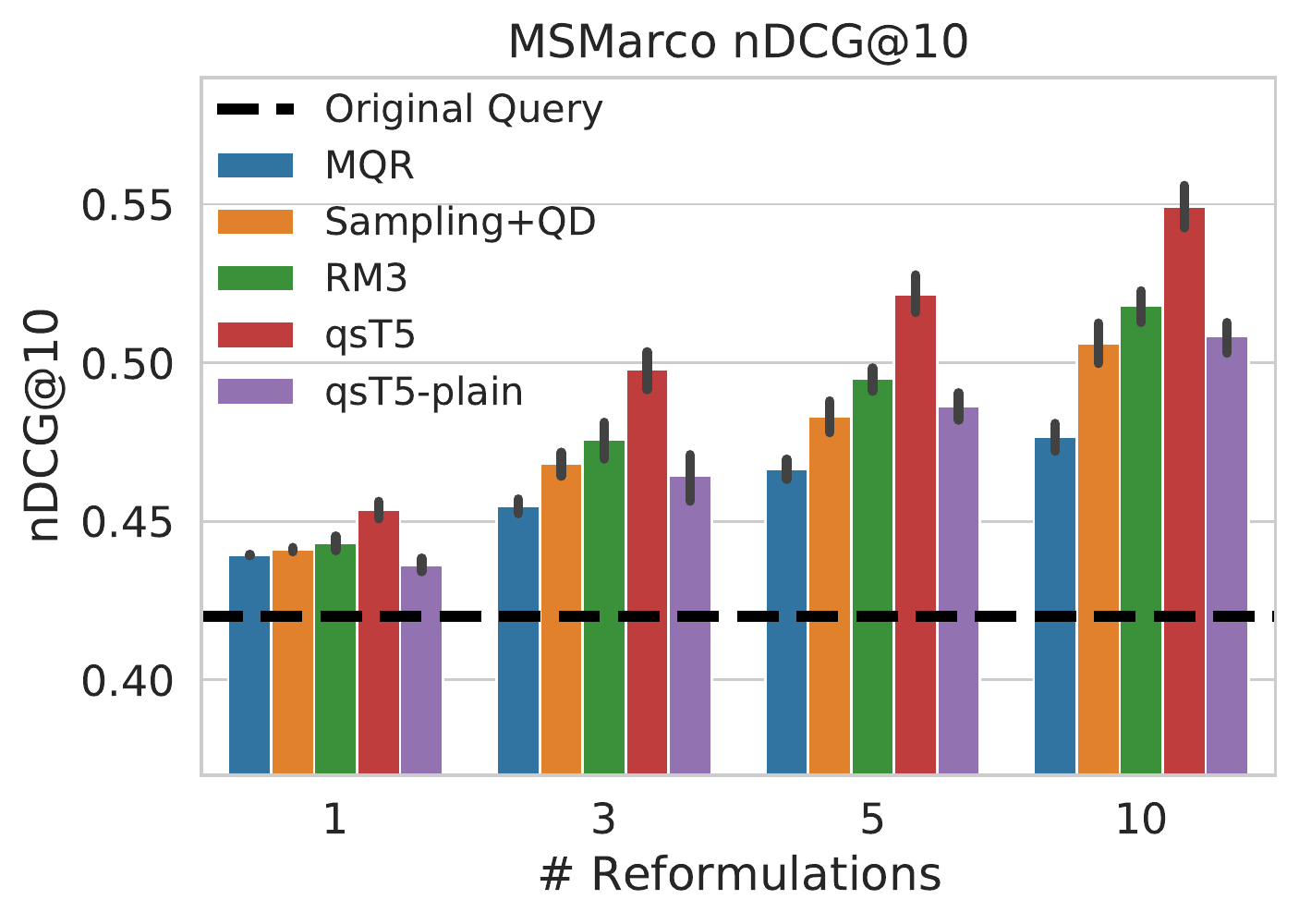}
        \caption{}
        \label{fig:reformulation_ndcg_msmarco}
    \end{subfigure}
    \begin{subfigure}[b]{.49\textwidth}
        \centering
        \includegraphics[width=1.\linewidth]{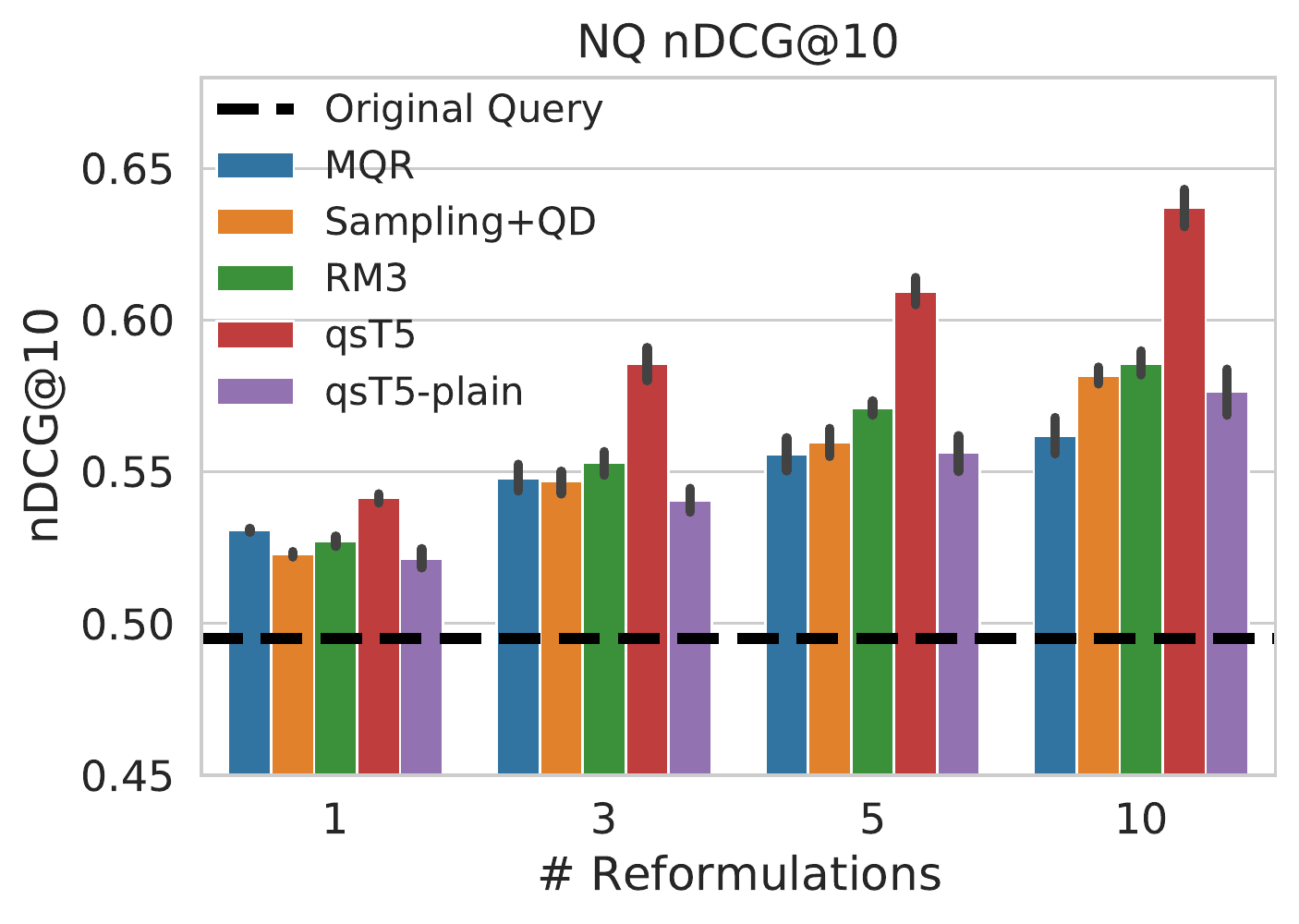}
        \caption{}
        \label{fig:reformulation_ndcg_nq}
    \end{subfigure}
    
    \caption{
     Retrieval metrics of the query suggestion models on the MSMarco (a) and NQ (b) test sets. The dashed line shows the nDCG@10 score of the original query. The bars represent the nDCG@10 of the best 1, 3, 5, and 10 reformulations (including the original query), respectively, of different models. The error bars show the 95 percent confidence interval when doing bootstrap sampling from up to 10 generations of the models. The RM3 and qsT5 models are using pseudo-relevance feedback, i.e., information about the top-retrieved paragraphs. 
    }
    \label{fig:reformulation_ndcg}
\end{figure*}

\renewcommand{\arraystretch}{1.2}
\begin{table}[t]
\centering
\resizebox{0.49\textwidth}{!}{%
    \begin{tabular}{@{}p{0.615\textwidth}@{}}
        \toprule
        \textbf{Original Query} \\
        \hspace{0.2cm} who created spiritual gangster \\ 

        \textbf{MQR} \\
         \hspace{0.2cm} Who created the Spiritual Gangster?  \\
         \hspace{0.2cm} Who created the ``spiritual gangster'' storyline? \\
         \hspace{0.2cm} Who created the ``spiritual gangster''?  \\

        \textbf{RM3} \\
        \hspace{0.2cm} who created spiritual gangster spiritual  \\
        \hspace{0.2cm} who created spiritual gangster modern  \\
         \hspace{0.2cm} who created spiritual gangster inspired  \\

        \textbf{Sampling+QD} \\
         \hspace{0.2cm} who created gangster a spiritual \& egantious  \\
         \hspace{0.2cm} who created spiritual gangster -gangster  \\
        \hspace{0.2cm}  who created spiritual gangster \\

        \textbf{qsT5} \\
         \hspace{0.2cm} who is the founder of spiritual gangsters  \\
         \hspace{0.2cm} who created the spiritual gangster ( spiritual yogi )  \\
         \hspace{0.2cm} what is the spiritual gangster movement \\

        \textbf{qsT5-plain} \\
        \hspace{0.2cm}  who are the founders of the gangster spirit band \\
         \hspace{0.2cm} how many gangsters were formed in white supreme \\
         \hspace{0.2cm} who was the members of the gangster supremes \\
        \midrule
        \textbf{Gold Paragraph} \\
        \hspace{0.2cm} About Spiritual Gangster. Spiritual Gangster represents a new generation of yogis seeking balance between the ancient practice of yoga and the modern world. Founded by Vanessa Lee and Ian Lopatin, this newly borne brand calls for high vibration living and radiating love shore-to-shore, person-to-person, heart-to-heart. \\ \bottomrule

    \end{tabular}%
}%
    \caption{Examples of the top query suggestions for the different models for the query ``who created spiritual gangster''. The final row shows the human-labeled gold paragraph.}
    \label{tab:ref_example_gangster}
\end{table}

\paragraph{Evaluation}
We evaluate the query suggestion models on the MSMarco and NQ test sets. For each example, we generate up to 10 suggestions using nucleus sampling \citep{nucleus_sampling}. Our ultimate goal is to provide users with at least one reformulation that better captures their search intent. As we assume the gold paragraph captures the information need of the user, we evaluate if, within a small set of reformulations, there is a query that would lead them closer to that paragraph; i.e., we measure the maximum nDCG@10 of the top-k reformulations and the original query.
We provide the results in Figure \ref{fig:reformulation_ndcg}. 

We see that our qsT5 model significantly outperforms all baselines on both datasets. Notably, it substantially improves upon the RM3 pseudo-relevance feedback baseline; this indicates that our full reformulation approach is more powerful for neural retrievers than a well-established query expansion technique.

The large gap between qsT5 and qsT5-plain validates the importance and usefulness of conditioning on the initial search results.

Successfully rewriting the query to a well-formed variant benefits this task as indicated by the improved nDCG performance of the non-PRF baseline of the T5 model trained on MQR (blue) over the original query (dashed line). The qsT5-plain model outperforms the MQR model when considering multiple reformulations on MSMarco, indicating that in some cases our model learns successful rewriting beyond improving fluency.

The qsT5-plain is mostly on par with sampling randomly around the embedding of the original query and using our query decoder to generate a reformulation; hence, we can speculate that the main benefit of this non-PRF model comes from an ensembling effect of generating suitable reformulations around the neighborhood of the original query. Again, this reinforces the benefit of pseudo-relevance feedback for the application of query suggestion.

Additional plots showing the inner product metric for this experiment and a table summarizing the numbers are provided in Appendix \ref{sec:additional_ref_results}.

\subparagraph{Diversity \& Fluency}
To quantitatively highlight the characteristics of the evaluated query suggestion models, we report Self-BLEU \citep{self_bleu} and perplexity of a language model as proxies for diversity and fluency, respectively, in Table~\ref{tab:bleu_ppl}. Self-BLEU is measured between 10 suggestions for a given query and averaged across the dataset, where a low Self-BLEU indicates large diversity between suggestions. For the perplexity evaluation, we employ the T5-base language model trained on C4 \citep{t5} and measure the average per-token perplexity of all suggestions for a given dataset; here, we associate lower perplexity with higher fluency of the suggestions.

Table \ref{tab:bleu_ppl} shows that the MQR baseline generates the most fluent queries, but with low diversity compared to our reformulation approaches. The RM3 query expansions score worst in diversity as they always use the original query as a base. Our qsT5 model scores second best in diversity, with a good comparative perplexity, only surpassed by the qsT5 variant without PRF, due to the fact that this model does not focus on the ``narrowed-down'' topics of the retrieved results. Notably, the perplexity is higher for the NQ dataset than for MSMarco due to the nature of queries in NQ being closer to well-formed questions as opposed to 'search engine queries'.

\renewcommand{\arraystretch}{1.1}
\begin{table}[ht]
    \centering
    \small
    \begin{tabular}{lrrrr}
    \toprule
         & \multicolumn{2}{c}{MSMarco} & \multicolumn{2}{c}{NQ} \\
        Model & S-BLEU & PPL & S-BLEU & PPL \\
        % \multirow{2}{*}{Model} & \multicolumn{2}{c}{MSMarco} & \multicolumn{2}{c}{NQ} \\
        % & S-BLEU & PPL & S-BLEU & PPL \\
        \midrule
        Original Query & - & 1622.2 & - & 217.9 \\
        MQR & 46.1 & 59.6 & 61.4 & 56.8 \\
        RM3 & 74.8 & 1562.6 & 88.0 & 309.5 \\
        Sampling+QD & 23.7 & 726.0 & 26.6 & 687.5 \\
        qsT5 & 17.8 & 247.8 & 18.4 & 223.2 \\
        qsT5-plain & 9.2 & 196.6 & 7.6 & 249.8 \\
    \bottomrule
    \end{tabular}
    \caption{Self-BLEU \citep{self_bleu} and Perplexity (PPL) for the query suggestions of the different models on MSMarco and NQ. Self-BLEU is measured between 10 suggestions of a model for a given query and then averaged across the dataset to provide information about the diversity of the suggestions. Perplexity is measured per token based on a T5 language model to provide a relative comparison of fluency of the query suggestions.}
    \label{tab:bleu_ppl}
\end{table}

\paragraph{Examples}
In Table~\ref{tab:ref_example_gangster}, we cherry-pick a representative example of query suggestions for the different models. This example showcases typical behaviors of the models. The MQR model is trained on turning ill-formed into well-formed questions. Hence, it usually produces grammatical but low diversity reformulations, especially when the original query is already close to a well-formed question. The relatively high Self-BLEU score amongst its reformulations for a given query, reported in Table~\ref{tab:bleu_ppl}, supports the argument of limited diversity.

The RM3 model appends the most relevant terms to the original query and therefore has the lowest overall diversity (i.e., highest Self-BLEU). The Sampling+QD model can result in non-grammatical or even nonsensical queries depending on the sampled point in latent space. While the qsT5 model can utilize the top-retrieved search results to form reformulations that are in accordance with the topic of the query (e.g., ``yogi'' in the example of Table~\ref{tab:ref_example_gangster}), the qsT5-plain needs to rely on its internal world knowledge stored in its parameters. It thus cannot connect ``gangster'' with ``yogi'' here.

\section{Conclusion} \label{sec:conclusion}
Dual encoders have reset the standard in IR. % and they are used primarily as black boxes.
However, language-based inverted index architectures still hold their ground, especially in out-of-domain evaluations~\citep{BEIR}.
To help further our understanding of the connections, and potential, between the two, we propose a method that relies on a query decoder to map back to language space the latent codes generated by the encoder. 

The interplay between latent and language representations, in combination with a simple goal-directed mechanism for traversing the shared query-document space, allows us to generate a large synthetic dataset of query reformulations on which we train a pseudo-relevance feedback query suggestion model that characteristically tries to predict the location of the target document.

Our contribution is twofold: (i) we develop a generic way to generate training data for directional query refinement by traversing the latent space between queries and relevant documents, and (ii) we build a powerful reformulation model that we evaluate on a novel benchmark inspired by the query suggestion task. Suggestions are typically well-formed, diverse and more likely to lead to the right document than competing methods.

\section{Limitations}
A proper user study would provide a valuable complement to the current evaluation and contribute to a fuller picture. However, this presents significant challenges that are beyond the scope of the current work. For instance, is not trivial to adequately design a meaningful task for human raters conducive to good agreement, e.g., it may be inevitable to second-guess the original query intent in the presence of unexpected interpretations brought to the surface by the suggestions. % and it would likely would unknown confounding factors, ambiguity, quality and agreement problems, etc. 
For the time being, we feel the automatic evaluation proposed here will be more valuable, as it makes direct comparison and reproducibility straightforward. 

Secondly, it seems sensible to further evaluate the query suggestions in an end-to-end IR task. Preliminary experiments in this direction using MS Marco proved somewhat inconclusive, while they introduce significant complexity. The data annotations are sparse (one passage per query, by and large) and it is often the case that multiple relevant passages exist for the same query.\footnote{Including near duplicate passages.} This makes reranking a crucial but faulty component, opening up a somewhat orthogonal front. The ideal evaluation would rely on a deeper manual analysis for a limited query set, e.g., TREC-style (e.g., cf.~\citet{trec-dl-2020}).

We leave both for future work.

\section*{Acknowledgements}
We would like to thank Léonard Hussenot for his insightful feedback and the support with the Diversity \& Fluency experiments. Furthermore, we thank Wojciech Gajewski, Sascha Rothe, and Lierni Sestorain Saralegui for helpful discussions and feedback throughout the course of the project.

\clearpage

\bibliography{anthology,custom}
\bibliographystyle{acl_natbib}

\clearpage

\onecolumn
\appendix
\section{Appendix}
\label{sec:appendix}

\subsection{Additional Results of the Query Suggestion Model}\label{sec:additional_ref_results}

\begin{table}[H]

\begin{tabular}{lllll|llll}
\toprule
          \multirow{2}{*}{\textbf{Model}} & \multicolumn{4}{c}{\textbf{MSMarco}} & \multicolumn{4}{c}{\textbf{NQ}} \\
           &     1&     3&     5&    10&          1&          3&          5&         10\\
\midrule
 Original Query &              .420 &                 - &                 - &                 - &              .495 &                 - &                 - &                 - \\
            MQR &  .439 \tiny{.001} &  .454 \tiny{.005} &  .464 \tiny{.005} &  .477 \tiny{.004} &  .531 \tiny{.001} &  .548 \tiny{.008} &  .557 \tiny{.009} &  .571 \tiny{.005} \\
    Sampling+QD &  .440 \tiny{.001} &  .469 \tiny{.005} &  .484 \tiny{.007} &  .506 \tiny{.013} &  .522 \tiny{.001} &  .548 \tiny{.005} &  .561 \tiny{.006} &  .580 \tiny{.005} \\
            RM3 &  .445 \tiny{.003} &  .472 \tiny{.016} &  .495 \tiny{.009} &  .522 \tiny{.011} &  .526 \tiny{.003} &  .552 \tiny{.012} &  .571 \tiny{.006} &  .589 \tiny{.012} \\
           qsT5 &  .455 \tiny{.002} &  .496 \tiny{.010} &  .519 \tiny{.010} &  .554 \tiny{.011} &  .541 \tiny{.003} &  .582 \tiny{.011} &  .615 \tiny{.008} &  .637 \tiny{.009} \\
     qsT5-plain &  .440 \tiny{.005} &  .470 \tiny{.007} &  .488 \tiny{.005} &  .508 \tiny{.008} &  .520 \tiny{.006} &  .543 \tiny{.006} &  .553 \tiny{.010} &  .577 \tiny{.013} \\
\bottomrule
\end{tabular}
    \caption{
     Retrieval metric nDCG@10 of the query suggestion models on the MSMarco and NQ test sets. The numbers represent the nDCG@10 of the best 1, 3, 5, and 10 reformulations (including the original query), respectively, of different models. The small number indicates the standard deviation  when doing bootstrap sampling from up to 10 generations of the models. 
     }
    \label{tab:my_label}
\end{table}

\begin{figure}[h]
    \centering
    \begin{subfigure}[b]{.49\textwidth}
        \centering
        \includegraphics[width=1.\linewidth]{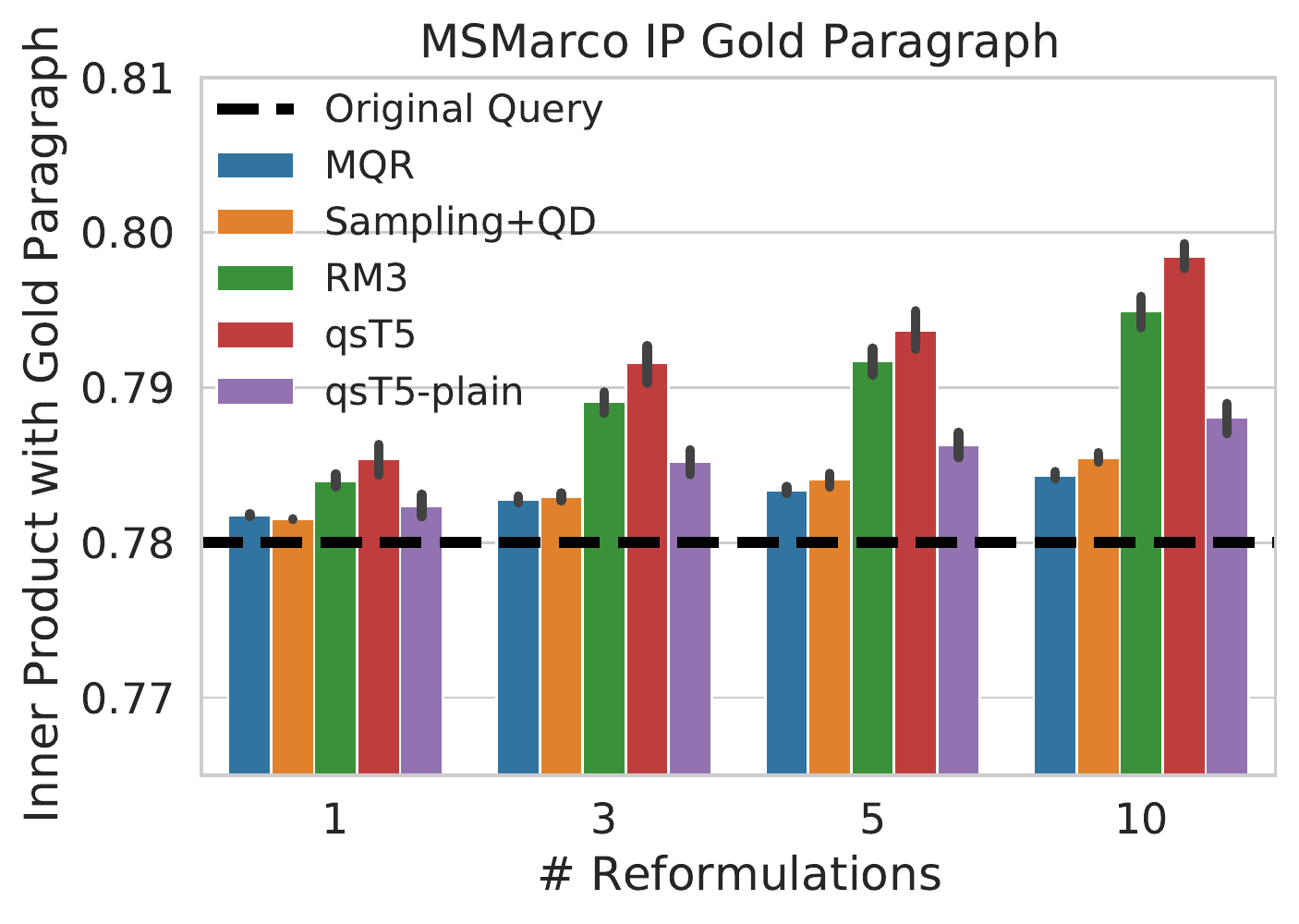}
        \caption{}
        \label{fig:reformulation_ip_msmarco}
    \end{subfigure}
    \begin{subfigure}[b]{.49\textwidth}
        \centering
        \includegraphics[width=1.\linewidth]{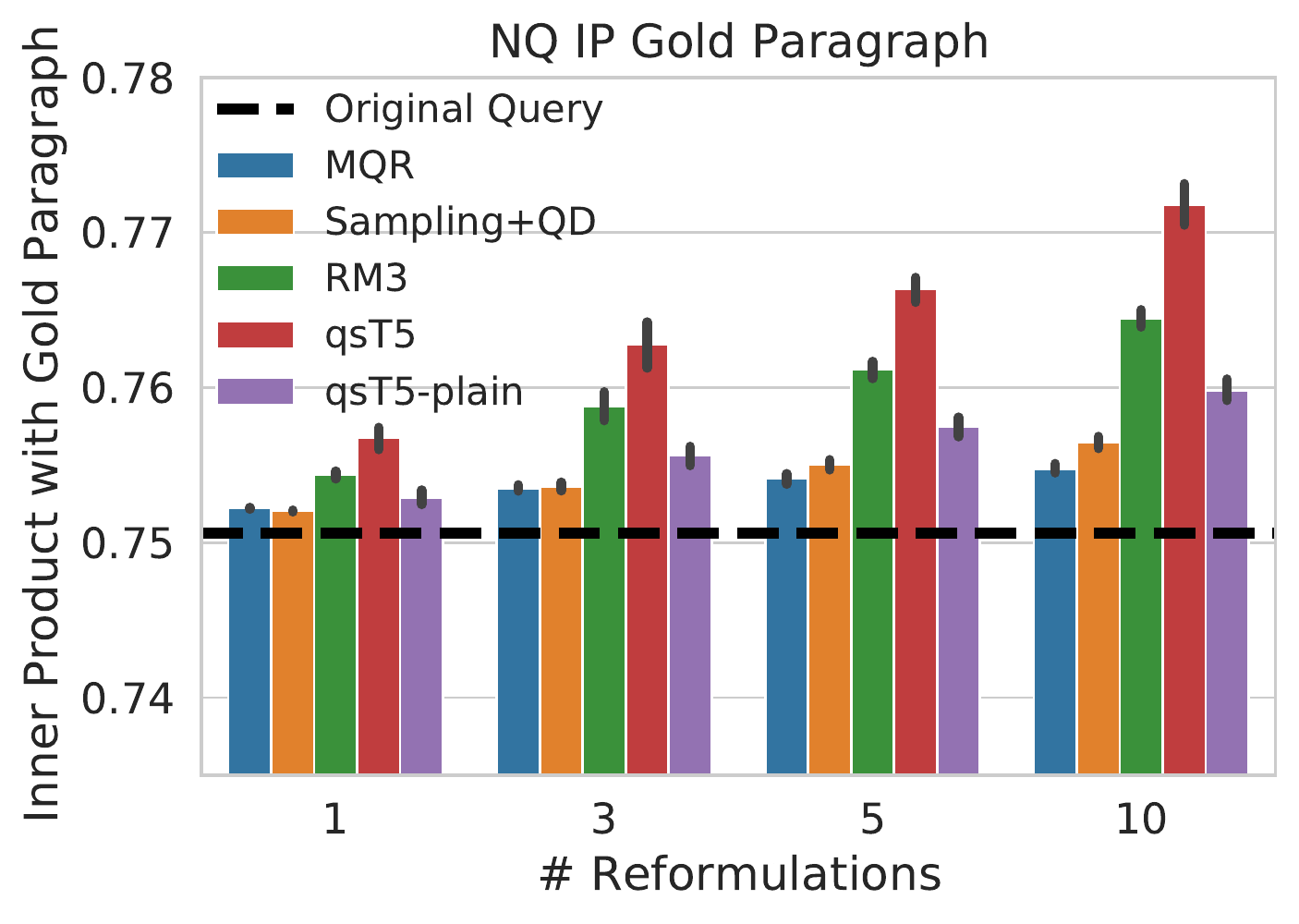}
        \caption{}
        \label{fig:reformulation_ip_nq}
    \end{subfigure}
    
    \caption{
     Inner product of the best reformulation with the gold paragraph for the various query suggestion models on the MSMarco (a) and NQ (b) test sets. The dashed line shows the inner product of the original query with the gold paragraph. The bars represent the inner product with the gold paragraph of the best 1, 3, 5, and 10 reformulations (including the original query), respectively, of different models. The error bars show the 95 percent confidence interval when sampling repeatedly from up to 10 generations of the models. The RM3 and qsT5 models are using pseudo-relevance feedback, i.e., information about the top-retrieved paragraphs, while the rest is only mapping from query to query.
    }
    \label{fig:reformulation_ip}
\end{figure}

\subsection{Training and Model Details}

\subsubsection{Query Decoder Model}\label{sec:qd_details}
We initialize the query decoder from the Tensorflow \citep{tensorflow2015-whitepaper} T5-base \citep{t5} checkpoint for conditional language generation of the Huggingface transformers library \citep{wolf-etal-2020-transformers}. We train and use only the decoder of the ~220M parameter model with 12 layers, 12 heads, and a hidden state dimension of 768. We train the model on a random sample of 3M queries from the PAQ dataset \citep{lewis-etal-2021-paq} with the hyperparameters provided in Table \ref{tab:qd_train_details} for ~130 hours on 16 Cloud TPU v3. Given the size of the model and the associated cost of training, we do not do an exhaustive grid search over the parameters but only sweep over four different values for the learning rate (1e-4, 3e-4, 5e-4, 1e-3). We determine the best model according to sequence classification accuracy on a held-out dev set.

\begin{table}[H]
\centering
\begin{tabular}{lr}
\toprule
\textbf{Parameter}  &\textbf{Value}  \\ \midrule
Number Decoder Layers & 12 \\
Number Heads & 12 \\
Head Dimension & 64 \\
Embedding Dimension & 768 \\
MLP Dimension & 2048 \\
\midrule
Batch Size & 64 \\
Dropout Rate & 0.0 \\
Base Learning Rate & 0.0005 \\
Warm-up Steps & 100 \\
Optimizer & AdamW\\
Weight Decay Rate & 0.01\\
Finetuning steps on PAQ & 5M \\
\bottomrule
\end{tabular}
\caption{Training Parameters for the T5-base Query Decoder model}
\label{tab:qd_train_details}
\end{table}

\subsubsection{qsT5 Query Suggestion Model}\label{sec:qst5_details}

We initialize our qsT5 and qsT5-plain models from the T5-large checkpoint and train them with the parameters provided in Table \ref{tab:qst5_train_details} for $\sim$24 hours on 16 Cloud TPU v3 each. We prefix the query with the keyword ``Query: '' and each passage, for the qsT5 model, with the keyword ``Paragraph: ''. For this model we rely on standard hyperparameters and, given the cost of training, do not further grid search for better values.

\begin{table}[H]
\centering
\begin{tabular}{lr}
\toprule
\textbf{Parameter}  &\textbf{Value}  \\ \midrule
Number Encoder Layers & 24 \\
Number Decoder Layers & 24 \\
Number Heads & 16 \\
Head Dimension & 64 \\
Embedding Dimension & 1024 \\
MLP Dimension & 2816 \\
\midrule
Batch Size & 128 \\
Dropout Rate & 0.1 \\
Base Learning Rate & 0.001 \\
Warm-up Steps & 1000 \\
Optimizer & AdaFactor \\
Input Token Length & 1024 \\
Output Token Length & 32 \\
\midrule
Finetuning steps (qsT5) & 40k \\
Finetuning steps (qsT5-plain) & 31k \\
\bottomrule
\end{tabular}
\caption{Training Parameters for the T5-large qsT5 and qsT5-plain Query Suggestion Models}
\label{tab:qst5_train_details}
\end{table}

\subsubsection{MQR Query Suggestion Model}\label{sec:mqr_details}
As with our qsT5 models, we initialize the MQR query reformulation model from the T5-large checkpoint. We train the model with the parameters provided in Table \ref{tab:mqr_train_details} for $\sim$3h on 16 Cloud TPU v3, and choose the best checkpoint according to sequence accuracy on the dev set. Training quickly overfits after about 2k step ($\sim$16min). We increased the batch size to 2048 and found the adding dropout didn't help but did no further hyperparameter tuning.

\begin{table}[H]
\centering
\begin{tabular}{lr}
\toprule
\textbf{Parameter}  &\textbf{Value}  \\ \midrule
Number Encoder Layers & 24 \\
Number Decoder Layers & 24 \\
Number Heads & 16 \\
Head Dimension & 64 \\
Embedding Dimension & 1024 \\
MLP Dimension & 2816 \\
\midrule
Batch Size & 2048 \\
Dropout Rate & 0.0 \\
Fixed Learning Rate & 0.001 \\
Optimizer & AdaFactor \\
Input Token Length & 32 \\
Output Token Length & 32 \\
\midrule
Finetuning steps & 1800 \\
\bottomrule
\end{tabular}
\caption{Training Parameters for the T5-large MQR Query Reformulation Model}
\label{tab:mqr_train_details}
\end{table}

\subsection{Dataset Details}\label{sec:dataset_details}
We experiment with a few different thresholds on what constitutes a successful reformulation for our generated dataset. We achieve the best results in terms of sequence classification accuracy of a held-out dev set of reformulations for the dataset described in the main body of the paper. Our final dataset of successful reformulations of MSMarco-train queries contains 863,207 successful query rewrites that are split among a training, development, and test set as reported in Table \ref{tab:dataset_details}.

\begin{table}[H]
\centering
\begin{tabular}{lr}
\toprule
\textbf{Split}  &\textbf{Number of Examples}  \\ \midrule
Train & 768,372 \\
Dev & 86,478 \\
Test & 8,457 \\
\midrule
Total & 863,307 \\
\bottomrule
\end{tabular}
\caption{Successful Reformulation Dataset Details}
\label{tab:dataset_details}
\end{table}

\subsection{Additional Reformulation Examples}\label{sec:additional_reformulation_examples}

\begin{table}[H]
    \small
    \centering
    \begin{tabular}{lp{8.5cm}r}
    
        \toprule
        \textbf{Type} & \textbf{Query} & \textbf{nDCG@10} \\\midrule
        
        Original Query & shopko kennewick address & 0.63 \\ \hline 
        
        MQR &  Is the address for the Shopko in Kennewick correct? & 1.00 \\
         & What is this shopko in Kennewick, NY address? & 1.00 \\
         & Is there a Shopko in Kennewick with the following address? & 1.00 \\
        \hline
        RM3 &  shopko kennewick address 867 & 0.50 \\
         & shopko kennewick address store & 1.00 \\
         & shopko kennewick address 5500 & 0.50 \\
        \hline
        Sampling+QD &  who is the localization of koieko shopphan & 0.00 \\
         & shopk kennewickko address in shopington & 0.63 \\
         & shopko kennewick is located in which address & 0.63 \\
        \hline
        qsT5 &  what is the address of shopko located in kennewick washington & 1.00 \\
         & what is the location of shopko in kennewick washington & 1.00 \\
         & where is shopko store in kennewick washington & 1.00 \\
        \hline
        qsT5-plain &  what time is open shopko at 867 kennewick & 1.00 \\
         & what is the main store at shopko kanetown & 0.00 \\
         & when does store for shopko located in north kansas & 0.00 \\
        \hline
        \hline 
        Gold Paragraph & \multicolumn{2}{p{13cm}}{Information about possible store closing and store hours for: ShopKo in Kennewick, Washington, 99336 Address: 867 North Columbia Center Blvd | Phone: (509) 736-0884 | Type: Store, Department Store, Retail More information:} \\ \bottomrule
    \end{tabular}
    \caption{Examples of the top query suggestions for the MSMarco query ``shopko kennewick address''. The final row shows the human-labeled gold paragraph and the nDCG@10 retrieval score is provided for each query. In this example, we see that the qsT5 model can leverage the PRF to successfully generate queries that focus on Kennewick in \textit{Washington}. For the non-PRF models, this is not possible.}
    \label{tab:ref_model_shopko}
\end{table}

\begin{table}[H]
    \small
    \centering
    \begin{tabular}{lp{8.5cm}r}
    
        \toprule
            \textbf{Type} & \textbf{Query} & \textbf{nDCG@10} \\\midrule
            
            Original Query & definition of a surge & 0.50 \\ \hline 
            
            MQR &  What is the definition of a surge? & 0.43 \\
             & What is the definition of a surge? & 0.43 \\
             & What is a surge? & 0.63 \\
            \hline
            RM3 &  definition of a surge surge & 0.33 \\
             & definition of a surge sudden & 0.43 \\
             & definition of a surge increase & 0.30 \\
            \hline
            Sampling+QD &  what is the definition of a surge & 0.50 \\
             & a a surge to a begin in the surge definition & 0.43 \\
             & what is the definition of a surge & 0.50 \\
            \hline
            qsT5 &  what is the definition of a surge in the sea & 0.50 \\
             & what is the meaning of surge in a wave & 0.63 \\
             & what is the definition of a surge in a sound & 1.00 \\
            \hline
            qsT5-plain &  what is the swiveling of a rolling motion & 0.00 \\
             & what is the meaning of a surge in the emotional side of a big wave & 0.43 \\
             & the surge of a wave or ayurvedic chorus & 0.00 \\
            \hline
            \hline 
            Gold Paragraph & \multicolumn{2}{p{13cm}}{surge. 1. a strong, wavelike forward movement, rush, or sweep: the surge of the crowd. 2. a sudden, strong rush or burst: a surge of energy. 3. a strong, swelling, wavelike volume or body of something. 4. the rolling swell of the sea. 5. a swelling wave; billow. 6. the swelling and rolling sea. 7. a. a sudden rush or burst of electric current or voltage. b. a violent oscillatory disturbance.} \\ \bottomrule

    \end{tabular}
    \caption{Examples of the top query suggestions for the MSMarco query ``definition of a surge''. The final row shows the human-labeled gold paragraph and the nDCG@10 retrieval score is provided for each query. Here, we see that the qsT5 model provides more useful additions to the query than the simple rephrasing of the MQR model. It adds topical terms like ``sea'', ``wave'', or ``sound'' with which it is possible to obtain improved retrieval performance.}
    \label{tab:ref_model_surge}
\end{table}

\begin{table}[H]
    \small
    \centering
    \begin{tabular}{lp{8.5cm}r}
    
            \toprule
            \textbf{Type} & \textbf{Query} & \textbf{nDCG@10} \\\midrule
            
            Original Query & what aircraft can you fly with a ppl & 0.29 \\ \hline 
            
            MQR &  What aircraft can you fly with a passenger? & 0.00 \\
             & What aircraft can you fly with passengers? & 0.00 \\
             & Which aircraft can you fly with a passenger? & 0.00 \\
            \hline
            RM3 &  what aircraft can you fly with a ppl pilot & 0.36 \\
             & what aircraft can you fly with a ppl license & 0.32 \\
             & what aircraft can you fly with a ppl private & 0.43 \\
            \hline
            Sampling+QD &  what aircraft can you fly with a ppl & 0.29 \\
             & what ships did you fly a learn park and can server ( pplalo & 0.00 \\
             & a able a flying wings able islands with a pistol pl & 0.00 \\
            \hline
            qsT5 &  what kind of airplane can you fly with a ppl & 0.00 \\
             & what type of aircraft are you able to fly a private pilot & 1.00 \\
             & what does a personal pilot ( a pvl ) mean & 0.00 \\
            \hline
            qsT5-plain &  what are the types of aircraft that can be used for private pilots & 1.00 \\
             & what kind of licenses are required to have a private pilot & 0.32 \\
             & how many pilots are required to take a commercial flight & 0.00 \\
            \hline
            \hline 
            Gold Paragraph & \multicolumn{2}{p{13cm}}{The types of aircraft one may fly depends on what they are certified for, e.g. Airplane Single Engine land. See Categories and Classes for a list. Usually, a newly minted private pilot is certified to fly all planes in the generic category single engine piston, often abbreviated SEP. This includes the Cessna 172, PA-28, Diamond DA40, Robin DR400 and similar planes. More complex types, like those with retractable undercarriage or variable pitch props, require additional training and licensing.} \\ \bottomrule

    \end{tabular}
    \caption{Examples of the top query suggestions for the MSMarco query ``what aircraft can you fly with a ppl''\footnotemark. The final row shows the human-labeled gold paragraph and the nDCG@10 retrieval score is provided for each query. In this example only the two qsT5 models are able to add the term ``private pilot'' to the query that leads to good retrieval performance. Notably, the RM3 PRF baseline ranks these two terms separately as important and adds them to their suggestions.}
    \label{tab:ref_model_nq_aircraft}
\end{table}
\footnotetext{The abbreviation PPL stands for \href{https://en.wikipedia.org/wiki/Private_pilot_licence}{private pilot license}.}

\subsection{Additional Traversal Examples}\label{sec:additional_traversal_examples}

\begin{figure*}[h]
    \centering
    \includegraphics[width=0.75\linewidth]{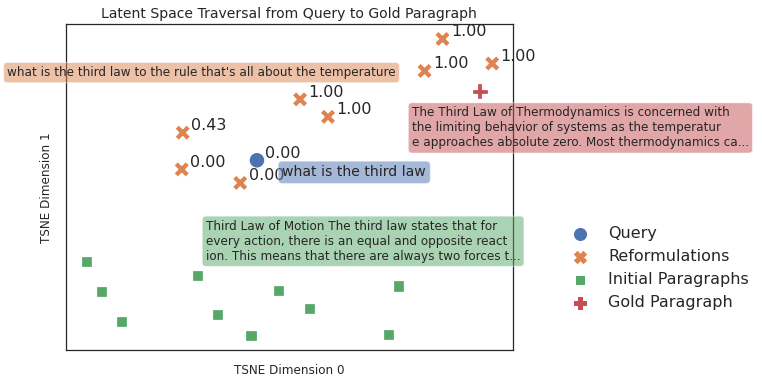}
    \caption{Visualization of the latent-space traversal from query to gold paragraph, using 2D t-SNE. The blue point denotes the embedding of the original query ``what is the third law''. The green squares are the embeddings of the retrieved paragraphs for this query. The closest one about the ``Third Law of Motion'' is shown in the green text bar. The orange ``x's'' denote the embeddings of the reformulations of the query decoder when moving to the gold paragraph depicted as the red plus. The orange and red text bars show the final reformulation and the gold paragraph text, respectively.}
    \label{fig:traversal_visualization_2}
    
\end{figure*}
\begin{table}[h]
    \centering
    \small
    \begin{tabular}{lp{10cm}r}
        \toprule
        Type & Query & nDCG@10 \\\midrule
        Original Query & where is quincy located & 0.00 \\ \hline 
        Reformulations &  where is quincy located in miami & 0.00 \\
         &  where is quincy located in a state & 0.00 \\
         & where is the city of quincy located & 0.00 \\
         &  where is the state of quincy located & 0.00 \\
         &  where is the city of quincy located in illinois & 0.39 \\
         &  where is the climate of quincy located in illinois & 1.00 \\
         &  where does the water come from in quincy illinois & 1.00 \\
         &  what is the average rainfall in quincy illinois & 1.00 \\
        \hline 
        Gold Paragraph & \multicolumn{2}{p{13cm}}{Location/Climate. The western most city in Illinois, Quincy is located along the eastern bank of the Mississippi River atop 90 foot limestone bluffs which overlook a wide expanse of the river and a natural harbor. Residents enjoy a moderate, four season climate where the sun shines nearly 68\% of the time. This area of Illinois receives an average of 36.86 inches of rainfall a year, and an average of 24 inches of snowfall. The average winter temperature is 28 degrees and the average summer temperature is 79 degrees.} \\
        \bottomrule
    \end{tabular}
    \caption{Example of a traversal on an MSMarco query. Here, we traverse the latent space in equidistant steps from the embedding of the original query ``where is quincy located'', which results in an nDCG@10 retrieval score of 0, to the embedding of the gold paragraph. The queries decoded from a latent code close to the gold paragraph are concerned with the climate of Quincy, Illinois, which is the focus of the gold paragraph.}
    \label{tab:traversal_visualization_session}
\end{table}

\begin{table}[h]
    \small
    \centering
    \begin{tabular}{lp{10.5cm}r}
        \toprule
        Type & Query & nDCG@10 \\\midrule
        
        Original Query & weather year round in new york city & 0.00 \\ \hline 
        
        Reformulations &  what is the weather year in new york city & 0.33 \\
         &  what is the weather in new york city each year & 0.00 \\
         &  what is the weather in new york city & 0.00 \\
         &  what is the hottest season in new york city & 0.39 \\
         &  when is the coldest month in new york & 0.30 \\
         &  when is the hottest month in new york & 1.00 \\
        \hline 
        Gold Paragraph & \multicolumn{2}{p{13cm}}{New York: Annual Weather Averages. July is the hottest month in New York with an average temperature of 25{\degree}C (76{\degree}F) and the coldest is January at 2{\degree}C (35{\degree}F) with the most daily sunshine hours at 11 in July. The wettest month is May with an average of 114mm of rain.} \\
        \bottomrule
    \end{tabular}
    \caption{Example of a traversal on an MSMarco query. Here, we traverse the latent space in equidistant steps from the embedding of the original query ``weather year round in new york city'', which results in an nDCG@10 retrieval score of 0, to the embedding of the gold paragraph. The queries decoded from a latent code close to the gold paragraph are more specific, asking about the hottest and coldest month in New York.}
    \label{tab:successful_traversal_nyc}
\end{table}

\begin{table}[h]
    \centering
    \small
    \begin{tabular}{lp{10cm}r}
        \toprule
        Type & Query & nDCG@10 \\
        \midrule
         Original Query &  what year declaration of independence & 0.00 \\
         \hline
         Reformulations &  when was the declaration of independence year & 0.00 \\
         &  when was the declaration of independence year & 0.00 \\
         &  what year is the declaration of independence & 0.00 \\
         &  when was the declaration of independence & 0.00 \\
         &  what is the declaration of independence in the year & 0.33 \\
         &  what is the declaration of independence in most of the person & 0.00 \\
         &  what is the declaration of independence in most of them & 0.00 \\
         &  when was the declaration of independence ( most important ) & 1.00 \\
         &  when was the declaration of independence ( dst ) on the most important document & 1.00 \\
         &  when was the declaration of independence ( most important ) & 1.00 \\
         \hline
         Gold Paragraph &  \multicolumn{2}{p{13cm}}{The Declaration of Independence is, of course, one of the country's most important documents, adopted at the Second Continental Congress on July 4, 1776. The text and purpose of the Declaration would likely be recognizable to those who have applied for U.S. citizenship, since questions about the document appear on the naturalization test.} \\
        \bottomrule
    \end{tabular}
    \caption{Example of a traversal on an MSMarco query. Here, we traverse the latent space in equidistant steps from the embedding of the original query ``what year declaration of independence'', which results in an nDCG@10 retrieval score of 0, to the embedding of the gold paragraph. The queries decoded from a latent code close to the gold paragraph, include the keyword ``important'' that seems to be helpful in retrieving this particular gold paragraph.}
    \label{tab:successful_traversal_independence}
\end{table}

% Example of retrieval which is good due to bad msmarco annotation (MSMarco fault).
\begin{table}[h]
    \centering
    \small
    \begin{tabular}{lp{10cm}r}
        \toprule
        Type & Query & nDCG@10 \\
        \midrule
         Original Query  &  how many us dollars are currently in circulation & 0.00 \\
         \hline
         Reformulations &  how many dollars are currently in circulation & 0.00 \\
             &  how many dollars are in circulation in us & 0.00 \\
             &  how many dollars are in circulation in the us & 0.00 \\
             &  how many dollars are in circulation in the us & 0.00 \\
             &  how many dollars are in circulation in the us & 0.00 \\
             &  what are the dollars that are in circulation us & 0.00 \\
             &  what are the sets of dollars in the us & 0.00 \\
             &  what are the denominations of dollars around us & 0.50 \\
             &  what are the denominations of \$ 100 were there & 1.00 \\
             &  what are the denominations of \$ more than that were used in the us & 1.00 \\
         \hline
         Gold Paragraph &  \multicolumn{2}{p{13cm}}{Large denominations of United States currency. Large denominations of United States currency greater than \$100 were circulated by the United States Treasury until 1969. Since then, the U.S. dollar has only been issued in seven denominations: \$1, \$2, \$5, \$10, \$20, \$50, and \$100.} \\
        \bottomrule
    \end{tabular}
    \caption{Example of a traversal on an MSMarco query. Here, we traverse the latent space in equidistant steps from the embedding of the original query ``how many us dollars are currently in circulation'', which results in an nDCG@10 retrieval score of 0, to the embedding of the gold paragraph. The decoded queries drift away from the original questions of dollars in circulation to the denominations of dollar bills. The reason of this drift becomes obvious by looking at the gold paragraph that explains the denominations of the US Dollar but not the money currently in circulation. This is one of many examples where the labeled gold paragraph is not correct and hence the decoded queries drift away from the original query.}
    \label{tab:msmarco_problem_traversal}
\end{table}

\begin{table}[h]
    \centering
    \small
    \begin{tabular}{lp{10cm}r}
        \toprule
        Type & Query & nDCG@10 \\
        \midrule
         Original Query  & are tesla electric cars & 0.00 \\
         \hline
         Reformulations & are tesla electric cars no are they electric car & 0.00 \\
             &  is tesla electric cars a car or they are electric & 0.00 \\
             &  tesla electric cars are they or electric car & 0.00 \\
             &  tesla electric cars are they being or electric & 0.00 \\
             &  tesla cars are electric to be or electric cars & 0.00 \\
             &  tesla electric cars are off of what kind of cars & 0.00 \\
             &  tesla electric cars are a bolton on which time & 0.00 \\
             &  tesla electric cars a mile off what speed are they & 0.00 \\
             &  737 el taton speedway was flying away on how many cars & 1.00 \\
             &  737 bolts on a ta speedway how many miles off airport & 1.00 \\
         \hline
         Gold Paragraph &  \multicolumn{2}{p{13cm}}{A Qantas Boeing 737 aircraft and a Tesla electric car race on the nearly 2 mile runway at Avalon Airport. The Tesla was hard to catch off the start. Both travelled neck and neck as the 737 reached take-off speed of 161 mph and the Tesla hit 155 mph. USA TODAY.} \\
        \bottomrule
    \end{tabular}
    \caption{Example of a traversal on an MSMarco query. Here, we traverse the latent space in equidistant steps from the embedding of the original query ``are tesla electric cars'', which results in an nDCG@10 retrieval score of 0, to the embedding of the gold paragraph. In this example the final decoded query perfectly solves the retrieval problem (nDCG of 1) even though being semantically very strange. These queries are an interesting example as their embedding is close to the embedding of the gold paragraph, even though they have little non-stopword overlap with the gold paragraph (in the second to last query, only ``737'' overlaps exactly).}
    \label{tab:msmarco_problem_traversal_2}
\end{table}

\end{document}